\documentclass[letterpaper, 10 pt, conference]{ieeeconf}  

\IEEEoverridecommandlockouts                              

\title{\LARGE \bf
Onboard Wind-Preview Model Predictive Control Using Pitot-Static Sensing for Multirotor UAVs
}

\usepackage{cite}
\usepackage{amsmath,amssymb,amsfonts}
\usepackage{graphicx}
\usepackage{textcomp}
\usepackage{xcolor}
\usepackage{booktabs}
\usepackage{multirow}
\usepackage{hyperref}
\usepackage{url}
\usepackage{caption}
\usepackage{subcaption}
\usepackage{algorithm}
\usepackage{algorithmicx}
\usepackage{epstopdf}
\usepackage{mathtools}
\usepackage{algpseudocode}

\author{Bas Meere$^{1}$, Eline Wisse$^{1}$, Laurens Vousten$^{1}$, Sander Doodeman$^{1}$, \\ Elena Torta$^{2}$, Paula Chanfreut$^{1}$ and Duarte Antunes$^{1}$
\thanks{*This research is supported by the Renewable Energy Transition Topsector Energy subsidies (HER+22-02-03430912) from the Dutch Ministry of Economic Affairs}%
\thanks{The authors are with the Department of Mechanical Engineering, Eindhoven University of Technology, 5612 AE Eindhoven, The Netherlands, in the $^{1}$Control Systems Technology group and the $^{2}$Robotics group. {\tt\small b.g.l.meere@tue.nl}}
}

\begin{document}

\maketitle
\thispagestyle{empty}
\pagestyle{empty}

\begin{abstract}
Effective wind gust rejection and stable hovering are critical for the outdoor operation of autonomous drones. However, existing gust rejection methods are primarily reactive, inferring the disturbance from the resulting motion or measuring it at the airframe. Either way, the wind has already begun to act before it can be compensated. In this work, we anticipate the gust instead by measuring the wind ahead of the drone with a low-cost, low-weight pitot-static sensor mounted on a boom. The resulting wind preview is incorporated into a nonlinear model predictive controller (MPC), which optimizes the drone motion while anticipating wind disturbances. A longer boom offers more preview time but adds inertia and degrades flight performance. We characterize this trade-off in simulation and show that the optimal preview distance is not a fixed property of the platform, but shifts with the wind speed and with how quickly the drone can respond. Indoor hardware experiments confirm the trend and show that the proposed controller substantially improves hover performance against a PX4 baseline and an otherwise identical wind-unaware MPC. Outdoor experiments show that the error along the wind direction is reduced by 54 percent with respect to the baseline, demonstrating that a single wind-aligned sensor can significantly improve hovering performance.
\end{abstract}

\section{INTRODUCTION}
Autonomous drones are being deployed for increasingly challenging and critical tasks such as infrastructure inspection\cite{lee2024DroneDrivenXRayImageBased, mendu2025StateoftheArtReviewApplication}, search-and-rescue operations \cite{lyu2023UnmannedAerialVehicles} and precision agriculture \cite{deoliveira2026SprayDepositionResponses}. A major challenge for these applications to be successful is that they require stable hovering flight even in windy conditions. In radiographic infrastructure inspection, for example, any disturbance during the X-ray exposure introduces motion blur that severely deteriorates the resulting image quality, and the required close proximity poses a significant safety risk to both the drone and the asset \cite{meere2025XrayImageGeneration, lee2025AutonomousDroneFlight, du2025DeblurringMethodologyMotionBlurred}. 

However, existing wind gust-rejection methods are predominantly reactive~\cite{arain2014RealtimeWindSpeed, OnboardFlowSensing, craig2020GeometricAttitudePosition}. Indirect approaches infer the disturbance from its effect on the drone motion, and thus require the wind to propagate through the system before it can be compensated \cite{soltaninezhad2025ReviewMethodsChallenges}. Direct approaches measure the wind locally, but with a sensor mounted near the drone center, they capture the flow only once it reaches the drone \cite{simon2023FlowDroneWindEstimation}. In both cases, the controller has to compensate for a disturbance that has already begun to act, a delay that bounds achievable performance and grows more limiting for larger, slower platforms. 

\begin{figure}[t]
    \centering
    \includegraphics[width=\linewidth]{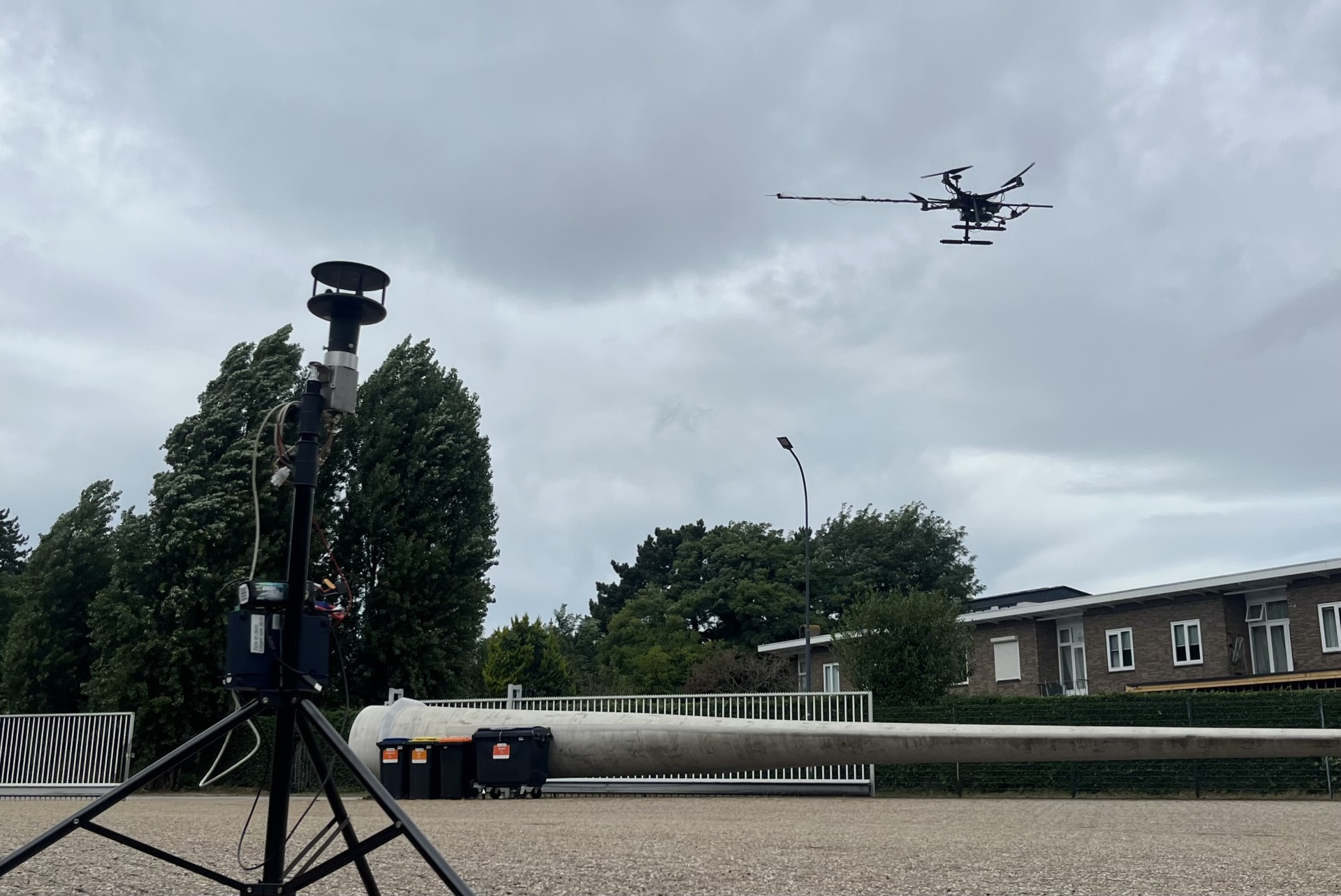}
    \caption{Outdoor flight experiment with a boom-mounted pitot-static sensor. The boom measures the wind upstream of the drone, giving the MPC a preview of incoming gusts to improve hover performance in windy conditions.}
    \label{fig:outdoor}
\end{figure}

Rather than reacting to wind disturbances after they occur, our proposed method therefore seeks to anticipate and counteract them by using \emph{preview} information. 
This requires two ingredients: a controller able to effectively use current and future disturbance information, and a means of sensing the wind ahead of the drone. In this work, the first is met by Model Predictive Control (MPC), which can include the known incoming disturbance directly into its prediction horizon and take preemptive action within the actuator limits~\cite{mendez2023WindPreviewBasedModel}. The second requires a sensor that can either measure at a distance or is light enough to be mounted at an offset from the drone.
Remote wind sensors such as Doppler LiDAR remain too heavy for a general multirotor, even in miniaturized form coming out at around 5 kg \cite{mendez2022ExperimentalVerificationLiDAR}. Keeping the LiDAR ground-based avoids this weight problem \cite{latif2025EnhancingQuadrotorResilience}, but ties the preview to a fixed location. Motivated by these shortcomings, in this article we mount a lightweight pitot tube on a boom extending ahead of the drone, providing direct spatial preview of the incoming, time-varying flow during flight. To the best of our knowledge, onboard wind preview sensing has not previously been combined with model predictive control on a multirotor. 

A consequence of this design is that it raises the question of how far ahead to measure. By increasing the boom length we can provide the quadrotor with more time to mitigate the incoming wind gust and thereby improve performance. 
However, this improvement is known to diminish and eventually saturate \cite{mendez2023WindPreviewBasedModel,khalil2021GustLoadAlleviation,dunne2013BenefitWindTurbine}. Furthermore, a longer boom affects flight performance, and wind measured further ahead has more time to decorrelate from the flow that ultimately reaches the drone. Therefore, alongside the preview control method itself, this work studies where the balance lies between the performance gained from earlier wind information and the cost of measuring it further from the drone, as well as the conditions on which that balance depends.

The main contributions of this work are summarized as follows:
\begin{itemize}
    \item We propose a nonlinear MPC running onboard a drone for preview-based gust rejection, using wind measurements obtained from a boom-mounted pitot-static sensor.  
    \item We characterize how controller performance depends on boom length, wind conditions and drone maneuverability when using a boom-mounted sensor for wind preview.
    \item We perform hover flight experiments in simulation and on hardware, both indoors and outdoors, that demonstrate the benefit of onboard wind preview measurements and validate the trade-off analysis. 
\end{itemize}
The remainder of this paper is organized as follows. Section \ref{Se:RW} reviews related work. Section~\ref{Se:GM} presents the boom-mounted pitot-static sensor and the filtering used to obtain a usable wind estimate. Section~\ref{Se:PC} formulates the MPC and the mapping from spatial preview to temporal horizon. Section~\ref{Se:Re} reports the boom length trade-off in simulation and the indoor and outdoor flight experiments, and Section~\ref{Se:Co} provides concluding remarks.


\section{RELATED WORKS}\label{Se:RW}
Drone-based wind-measurement is a well established field spanning modalities from pressure probes to hotwire, ultrasonic, and LiDAR sensors \cite{soltaninezhad2025ReviewMethodsChallenges,abichandani2020WindMeasurementSimulation}. However, obtaining these measurements at a preview is more challenging, as it requires the sensor to either be able to measure at a distance from the drone itself or be light enough to be mounted away from the center of the drone without significantly degrading flight performance. Wind LiDARs are the most established remote sensor for this purpose, but their weight and size make them unsuitable for multirotor applications, especially when wind sensing is a supplementary sensor rather than the main payload. This limits them to larger aircraft \cite{yew2025ReviewAirborneDopplerb, fezans2017InflightRemoteSensing} or ground-based \cite{latif2025EnhancingQuadrotorResilience, sinner2022ExperimentalTestingPreviewEnabled} deployment, where ground-based sensing additionally suffers from limited spatial coverage and local obstructions. To obtain preview, therefore, we opt for onboard sensing with a lightweight sensor attached to the drone via a boom.

Among onboard airflow sensors, 2D ultrasonic anemometers and pitot tubes offer the lightest designs and both have seen boom-mounted applications \cite{arain2014RealtimeWindSpeed, fuertes2019MultirotorUAVBasedPlatform, brewer2020MeteorologicalProfilingFire}. Ultrasonic units resolve the full in-plane wind vector and remain accurate down to near-zero speeds, but even the lightest units weigh 50 g or more and update at tens of hertz \cite{basawanal2025ExperimentalStudyUltrasonic}. At roughly 5 g and several hundred hertz, a pitot tube is an order of magnitude lighter and faster, at a fraction of the cost. Its slender form also mounts inline along the boom with little flow interference. Furthermore, as preview is most beneficial when the boom is oriented into the wind, the relevant quantity is mainly the wind component along the boom axis, so the benefit of the multi-axis capability of an ultrasonic sensor is limited here. Pitot tubes are consequently a strong fit as a low-cost, low-weight solution for obtaining preview information. 

However, obtaining the measurement is not enough as the controller must also be able to exploit it. Various methods use direct wind measurements to improve flight performance, through feedforward compensation \cite{arain2014RealtimeWindSpeed}, feedback linearization \cite{OnboardFlowSensing, craig2020GeometricAttitudePosition} and learned residual policies \cite{simon2023FlowDroneWindEstimation}, but all sense the flow at or near the airframe, so the gust has already begun to act by the time it is measured. Arain et al.~\cite{arain2014RealtimeWindSpeed} come closest to our setup with a boom-mounted pitot-static sensor, but the offset is mostly used to clear the rotor downwash and the compensation responds to the instantaneous reading regardless of where it was taken. Methods that do exploit future wind disturbance information include preview feedforward \cite{hamada2023RobustPreviewFeedforward}, preview $\mathcal{H}_\infty$ \cite{khalil2021GustLoadAlleviation} and MPC. MPC is particularly well suited, as it admits the previewed wind directly into its prediction horizon while enforcing state and input constraints, and it has been applied extensively to wind preview control of wind turbine blades ~\cite{sinner2022ExperimentalTestingPreviewEnabled, arkhouch2026WindSpeedForecasting}. In UAV-based deployment, Mendez et al.~\cite{mendez2023WindPreviewBasedModel} and Latif et al.~\cite{latif2025EnhancingQuadrotorResilience} embed LiDAR preview in an MPC and explicitly account for the transport delay to the drone. They show that wind preview retains its benefit even under sizeable timing (1.75 s) or magnitude (120\%) errors. However, their work relies on a ground-based sensor and they demonstrate flight performance only in simulation. Wang et al. \cite{wang2015ExperimentalVerificationModel} also use MPC, but obtain the preview from an earlier pass along the same path. This requires the disturbance to be steady and repeatable, which does not hold for dynamic gusts. To the best of our knowledge, no prior work combines onboard preview wind measurement with MPC on a multirotor and demonstrates it in real-world deployment. 


\section{WIND SENSING}\label{Se:GM}
\begin{figure}[t]
    \centering
    \includegraphics[width=0.9\linewidth]{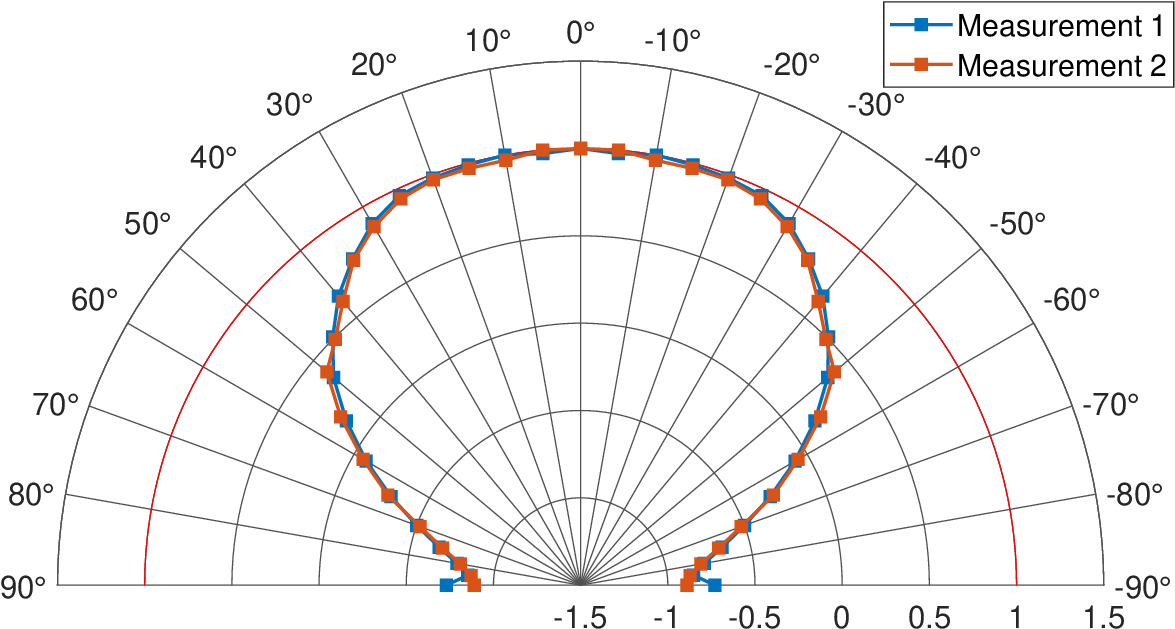}
    \caption{Angle sensitivity of the pitot-static sensor. A pressure coefficient $C_p$ = 1 means the probe registers the full flow. Two repeated sweeps are shown.}
    \label{fig:cp}
\end{figure}

In our work, wind speed is measured using a boom-mounted single-hole pitot-static sensor. It is offset from the center of gravity and aligned with the body longitudinal $x$-axis so it samples the flow ahead of the airframe (Fig.~\ref{fig:outdoor}). The probe returns a scalar along-axis airspeed obtained from the dynamic pressure, $V_p=\sqrt{2(p_t-p_s)/\rho}$, with $p_t$, $p_s$ the total and static pressures and $\rho$ the air density. A single-hole probe is directionally sensitive: the reading peaks with the flow aligned to the boom and decays with increased angle $\alpha$. This sensitivity can be characterized with the pressure coefficient \cite{beck2010AerodynamicsPitotStatic}
\begin{equation}\label{eq:cp_correction}
    C_p(\alpha) = \frac{p_t - p_s}{\tfrac{1}{2}\rho V^2}, \qquad C_p(0)=1 ,
\end{equation}
where $V$ is the true airspeed. As shown in Fig.~\ref{fig:cp}, $C_p$ stays essentially flat within a $\pm20^\circ$
acceptance cone and falls off beyond it. When the drone tilts to reject a gust, the boom axis might leave this cone relative to the horizontal wind. Using the tilt angle from the onboard attitude estimate, we invert $C_p(\alpha)$ to obtain the corrected airspeed $v^w=V_p/\sqrt{C_p(\alpha)}$. To get the most benefit of the preview with a single sensor requires that the boom is oriented toward the main direction of the wind. This should either be known in advance or can be estimated by letting the drone slightly drift. The boom allows us to test at different preview distances although a minimum distance is required to prevent adverse effects of the rotor downwash on the pitot readings. 

\begin{figure}[t]
    \centering
    \includegraphics[width=\linewidth]{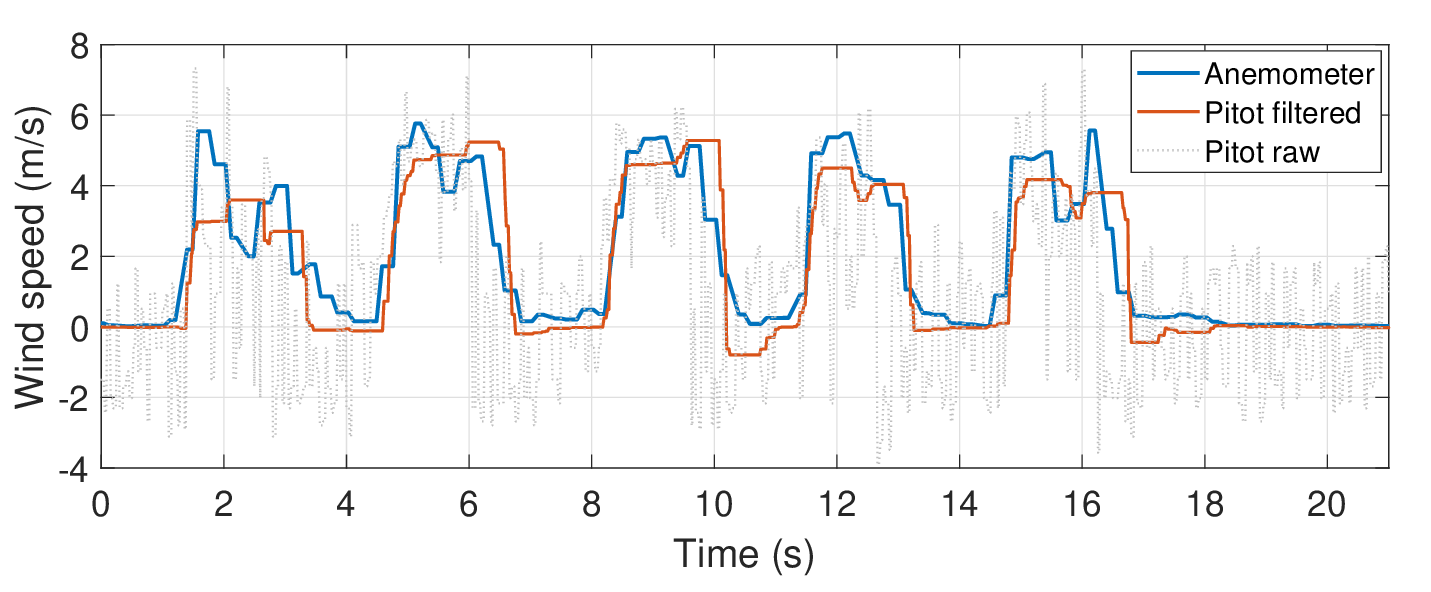}
    \caption{Wind speed measurement comparison between a static anemometer and the pitot tube mounted on a hovering drone. Even at low wind speed and while hovering the pitot sensor is able to capture the wind speed effectively. }
    \label{fig:pit_vs_ane}
\end{figure}

The raw readings pass a three-stage filtering pipeline, each stage targeting one error mode. First, Gaussian noise caused by drone vibrations is suppressed using an Exponential Moving Average (EMA). Second, a short rolling maximum rejects the brief dips that we observed due to off-axis turbulence. Third, readings below roughly 1 m/s were found to be unreliable for our sensor. This is mainly because at low differential pressures, sensor noise and ADC quantization translate into an increasingly large airspeed uncertainty through the square-root pressure–velocity relationship. Rather than a hard cutoff, a sigmoid function centered at $V_\mathrm{th}=1$~m/s smoothly down-weights measurements as they enter this low-confidence regime, avoiding discontinuities in the reconstructed signal. 

Figure~\ref{fig:pit_vs_ane} compares a stationary ultrasonic anemometer (assumed ground truth) with a pitot measurement captured while hovering in our indoor experimental setup. The filtered estimate tracks the reference well with an RMSE of 0.72 m/s over the interval shown. Although the achieved RMSE is higher than that typically reported for LiDAR-based wind sensing or ultrasonic anemometers \cite{mendez2022ExperimentalVerificationLiDAR, brewer2020MeteorologicalProfilingFire}, the pitot probe provides a compelling trade-off between accuracy, mass, cost, and deployability. Weighing only a few grams and requiring no external infrastructure, the probe captures even low-speed gust dynamics with sufficient accuracy for control purposes (as will be described in Section~\ref{Se:Re}), making it well suited as an onboard preview sensor.


\section{WIND PREVIEW CONTROL} \label{Se:PC}
\begin{figure}[t]
    \centering
    \includegraphics[width=\linewidth]{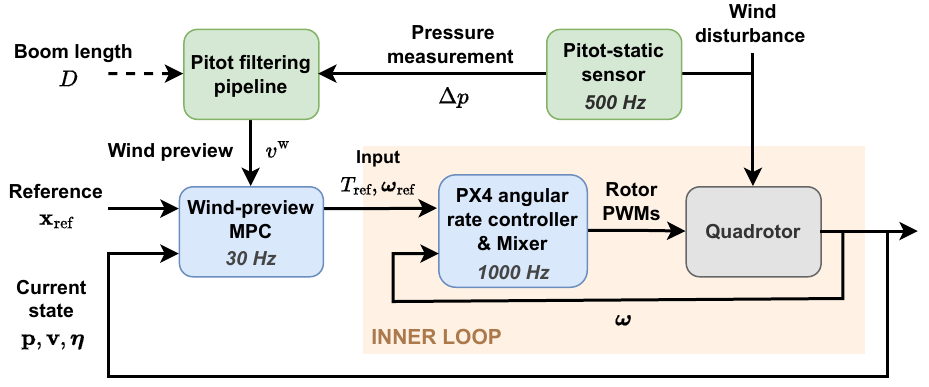}
    \caption{Control diagram of the wind-preview MPC including wind sensing and the PX4 inner loop. The boom length $D$ is shown dashed as it is a configuration parameter rather than a runtime signal.}
    \label{fig:MPCarchitecture}
\end{figure}

In this section, we discuss the formulation of an MPC that uses the preview information in its horizon to generate anticipatory actions. An overview of the control architecture is shown in Fig. \ref{fig:MPCarchitecture}. 

\subsection{System dynamics}\label{sec:rom}
Consider a quadrotor operating in an inertial North-East-Down (NED) frame $\mathcal{I} = \{\mathbf{e}_1^i, \mathbf{e}_2^i, \mathbf{e}_3^i\}$, with a body-fixed Forward-Right-Down (FRD) frame $\mathcal{B} = \{\mathbf{e}_1^b, \mathbf{e}_2^b, \mathbf{e}_3^b\}$ at the centre of mass. The position $\mathbf{p} = [x, y, z]^\top$ and velocity $\mathbf{v} = [v_x, v_y, v_z]^\top$ are expressed in $\mathcal{I}$. The attitude $\boldsymbol{\eta} = [\phi, \theta, \psi]^\top$ gives the ZYX Euler angles of $\mathcal{B}$ relative to $\mathcal{I}$, and $\boldsymbol{\omega} = [\omega_x, \omega_y, \omega_z]^\top$ is the angular rate in $\mathcal{B}$. The state and input vectors are $ \mathbf{x} = [\mathbf{p}^\top,\, \mathbf{v}^\top,\, \boldsymbol{\eta}^\top]^\top$ and $\mathbf{u} = [T,\, \boldsymbol{\omega}^\top]^\top$. 

We use a cascaded control structure where the MPC commands collective thrust and body rates $\mathbf{u}$, while the PX4 inner loop, running at a much higher rate, tracks $\boldsymbol{\omega}$. Due to this difference in rate, we treat the rate tracking as instantaneous and omit the rotational dynamics, inertia tensor, and motor mixing from the prediction model. This yields the quadrotor dynamics as, 
\begin{equation}\label{eq:rom_dynamics}
\begin{aligned}
    \dot{\mathbf{p}} &= \mathbf{v},\\
    m\dot{\mathbf{v}} &= m g\, \mathbf{e}_3^i - T\, R(\boldsymbol{\eta})\, \mathbf{e}_3^b
        + \mathbf{F}_A,\\
    \dot{\boldsymbol{\eta}} &= W(\boldsymbol{\eta})^{-1}\boldsymbol{\omega},
\end{aligned}
\end{equation}
where $m$ is the mass, $g$ is the gravitational constant and $T$ is the total thrust. The rotation matrix from $\mathcal{B}$ to $\mathcal{I}$ follows the ZYX convention,
$R(\boldsymbol{\eta}) = R_z(\psi)\,R_y(\theta)\,R_x(\phi)$, and $W(\boldsymbol{\eta})^{-1}$ maps the body rates $\boldsymbol{\omega}$ to the
Euler-angle rates $\dot{\boldsymbol{\eta}}$. The explicit entries of $R$ and $W$ follow the standard ZYX forms~\cite{diebelRepresentingAttitudeEuler}. The aerodynamic force $\mathbf{F}_A$ couples the wind into the translational dynamics. It is expressed in $\mathcal{I}$ as
\begin{equation}\label{eq:drag}
    \mathbf{F}_A = -\tfrac{1}{2}\,\rho\, C_d A\,
    \lVert \mathbf{v} - \mathbf{v}^\mathrm{w} \rVert\,(\mathbf{v} - \mathbf{v}^\mathrm{w}),
\end{equation}
where $C_d$ is the drag coefficient and $A$ is the effective frontal area~\cite{tran2015QuadrotorControlWind}. The sensor is aligned with the body $x$-axis and the boom is oriented toward the wind as described in Sec.~\ref{Se:GM}, so the drone heading $\psi$ coincides with the wind direction. The previewed wind in $\mathcal{I}$ is then reconstructed by a rotation about the vertical axis,
\begin{equation}\label{eq:wind_recon}
    \mathbf{v}^\mathrm{w} = v^\mathrm{w}\, R_z(\psi)\, \mathbf{e}_1^i.
\end{equation}

\subsection{Controller design}
The continuous-time dynamics~\eqref{eq:rom_dynamics} are discretized via fourth-order Runge-Kutta at sampling time $T_s$, giving
\begin{equation}\label{eq:discrete_dynamics}
    \mathbf{x}_{j+1} = f(\mathbf{x}_{j},\mathbf{u}_{j},\mathbf{v}^\mathrm{w}_{j}),
\end{equation}
with $\mathbf{v}^\mathrm{w}_{j}$ the previewed wind at step $j$.
Based on this model, we solve the following optimal control problem at each time step $k \in \mathbb{N}$:
\begin{alignat}{2}
    \min_{X_k,U_k}  & \sum_{j=0}^{N-1} (\lVert\mathbf{x}_{j|k} - \mathbf{x}_\mathrm{ref} \rVert_{Q}^{2} + \lVert\mathbf{u}_{j|k}\rVert^{2}_{R})
     + \lVert\mathbf{x}_{N|k} - \mathbf{x}_\mathrm{ref}\rVert_{P}^{2}, \notag \\
    \text{s.t.} \quad & \mathbf{x}_{j+1|k} = f(\mathbf{x}_{j|k},\mathbf{u}_{j|k}, \mathbf{v}^\mathrm{w}_{j|k}), \quad  j \in \{0, \dots, N-1\}, \notag \\
    & \mathbf{x}_{0|k} = \mathbf{x}_k, \notag \\
    & \underline{\mathbf{x}} \leq \mathbf{x}_{j|k} \leq \overline{\mathbf{x}}, \qquad j \in \{0, \dots, N\}, \notag \\
    & \underline{\mathbf{u}} \leq \mathbf{u}_{j|k} \leq \overline{\mathbf{u}}, \qquad j \in \{0, \dots, N-1\}, 
    \label{eq:NonlinearMPC}
\end{alignat}
where $\mathbf{x}_{j|k}$ and $\mathbf{u}_{j|k}$ denote the predicted state and input 
$j$ steps ahead of the current controller time step $k$. The reference state is $\mathbf{x}_\mathrm{ref}$, and $Q \succeq 0$, $R \succ 0$ and $P \succeq 0$ are diagonal weighting matrices penalizing the state error, the control effort and the terminal state error, respectively. The finite prediction horizon is $N \in \mathbb{N}_{\geq1}$. Accordingly, the sequence of current and predicted states $X_k = [\mathbf{x}_{0|k}, \dots, \mathbf{x}_{N|k}]$ depends on the discretized system dynamics $f(\cdot)$, the initial condition $\mathbf{x}_{0|k}$, and the sequence of current and predicted control inputs $U_k = [\mathbf{u}_{0|k}, \dots, \mathbf{u}_{N-1|k}]$. The lower and upper bounds on the states and inputs are denoted by $\underline{\mathbf{x}}$, $\overline{\mathbf{x}}$, $\underline{\mathbf{u}}$ and $\overline{\mathbf{u}}$, respectively. 

The sensor is mounted at a fixed distance from the drone and as such provides a spatial preview, whereas the MPC horizon is a temporal lookahead. To map one to the other, control architectures commonly rely on the Taylor Frozen Turbulence Hypothesis~\cite{taylor1938SpectrumTurbulence,dunne2014ImportanceLidarMeasurement, mendez2022ExperimentalVerificationLiDAR}, which assumes the turbulent wind field is ``frozen" in time and progresses downstream at a constant mean velocity. Based on this, the time $\tau$ that it takes for the wind to cross the sensor-to-drone distance $D$, i.e. the boom length, is defined as
\begin{equation}\label{eq:timedelay}
    \tau = \frac{D}{V_{\mathrm{safe}}}, \quad V_{\mathrm{safe}} = \max\!\left(\mathrm{MA}(v^{\mathrm{pitot}}),\, V_{\mathrm{th}}\right),
\end{equation}
where $\mathrm{MA}(\cdot)$ denotes a moving average of the filtered wind speed over a window $T_{\mathrm{MA}}$. The window is chosen to be long relative to the gust timescale. This is distinct from the EMA of Section~\ref{Se:GM}, which suppresses vibration noise in the instantaneous signal. The wind speed is lower bounded by $V_\mathrm{th}$ to keep $\tau$ finite at near-zero wind.

The pitot is sampled much faster than the control rate, and its readings are stored in a timestamped buffer $B = \{(t_i, v^{\mathrm{pitot}}_i)\}$. The predicted time at horizon step $j$ of controller step $k$ is $t_{j|k} = t_k + j\,T_s$, where $t_k$ is the timestamp of step $k$. The previewed airspeed at that step is recovered by linearly interpolating the buffer at the arrival time $t_{j|k} - \tau$,
\begin{equation}\label{eq:preview}
    v^\mathrm{w}_{j|k} =
    \begin{cases}
        \mathrm{interp}\!\left(B,\; t_{j|k}-\tau\right), & j\,T_s < \tau,\\[2pt]
        v^\mathrm{w}_{j-1|k}, & j\,T_s \geq \tau,
    \end{cases}
\end{equation}
for $j \in \{0,1,\dots,N\}$, where $\mathrm{interp}(B,t)$ linearly interpolates the buffer between the two readings closest to time $t$. The second case applies when the horizon extends beyond the available preview, in which case the last measured value is held constant.

\section{RESULTS} \label{Se:Re}
In this section we evaluate the proposed wind-preview MPC in simulation and on hardware. In simulation we assess the effect of the boom length across different wind conditions and platform response speeds. Indoors we compare against two baseline controllers and confirm the boom length trend over the range that can be flown safely. Outdoors we demonstrate performance in real wind.

\subsection{Experimental setup}
The platform used in this work is a Holybro S500 quadrotor with a Pixhawk 6C flight controller. A Holybro M10 GPS module and an OptiTrack motion capture system provide position feedback for outdoor and indoor flights, respectively. Wind is measured with a Mateksys AS-DLVR-I2C digital airspeed sensor mounted on a carbon boom, which gives a rigid mounting at low added mass. The controller runs onboard on a Raspberry Pi 4 companion computer. The proposed algorithm is implemented in Python and solved with Acados \cite{verschueren2022AcadosModularOpensource} in a receding horizon scheme using SQP-RTI. The MPC outputs body rates and a collective thrust, sent to PX4 in offboard mode over uXRCE-DDS, where the onboard rate controller tracks them. The MPC runs at 30 Hz, with a solve time on the order of 10 ms on hardware against approximately 4 ms in simulation (Intel i7-11800H). When not explicitly mentioned otherwise, we use a 1 meter boom on the drone. Parameters common to the simulation and hardware experiments are listed in Table~\ref{tab:parameters}.


\begin{table}[t]
    \centering
    \caption{Parameters used in simulation and on hardware.}
    \begin{tabular}{ll|ll} \toprule
        \textbf{Parameter} & \textbf{Value} & \textbf{Parameter} & \textbf{Value} \\ \midrule
        $m_\mathrm{drone}$    & 2.06 kg   &$Q$            & diag(5,5,3,1,1,1,6,6,6)\\
        $\mu_\mathrm{boom}$   & 41 g $\text{m}^{-1}$    & $R$              & diag(0.1,1,1,1) \\
         $m_\mathrm{sensor}$     & 5 g  & $P$              & 2$Q$ \\
         $A$   & 0.1 $\text{m}^2$ & $N$              & 30 \\ 
          $C_d$   & 0.5 & $T_\mathrm{s}$      & 1/30 s  \\
        $\alpha_\mathrm{EMA}$ & 0.05      & $T_{\mathrm{MA}}$         & 60 s (0.1 s indoor) \\
          $V_\mathrm{th}$ & 1 m $\text{s}^{-1}$  & $T_{\mathrm{RM}}$         & 0.1 s     \\
        \bottomrule
    \end{tabular}
    \label{tab:parameters}
\end{table}

\subsection{Simulation results}
\begin{figure}[t]
    \centering
    \includegraphics[width=\linewidth]{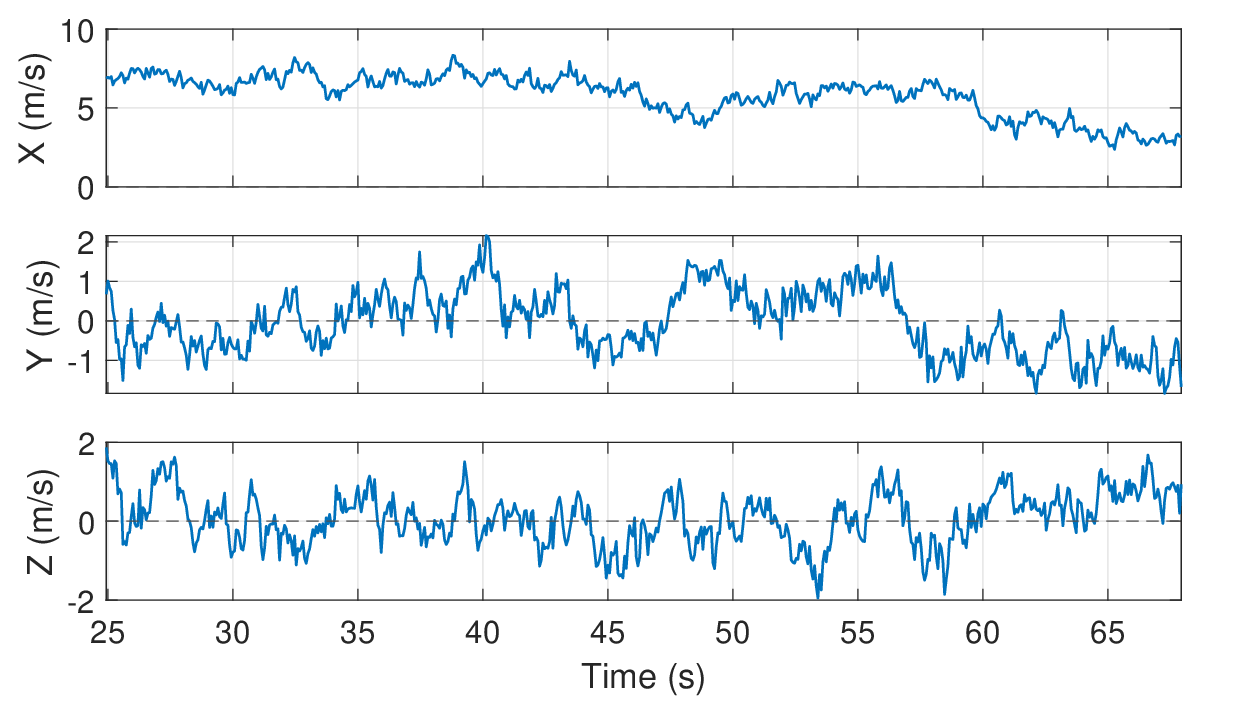}
    \caption{Segment of longitudinal, lateral and vertical wind components of a 7 m/s field generated under IEC class A normal turbulence. Note the different vertical scales.}
    \label{fig:wind}
\end{figure}

We evaluate the impact of boom length on tracking performance in simulation across a range of wind conditions. Simulations are performed in Gazebo using the proposed wind-preview MPC and PX4 software-in-the-loop, so the offboard interface and inner-loop rate controller are the same as on hardware. The wind fields are generated with evoTurb \cite{chen2022FourdimensionalWindField}, a generator originally developed for simulating LiDAR-based wind preview control of wind turbines, which includes wind evolution along the wind direction. The underlying turbulence is produced by TurbSim under IEC class A normal turbulence (NTM), from which we generate four mean wind conditions: 2, 4, 7 and 10 m/s. The 7 m/s field is visualized in Fig.~\ref{fig:wind}. Each test consists of a 70 s hover at 2 m altitude, repeated 10 times. The drone inertia is increased with boom length according to the linear mass density of the boom and the sensor mass given in Table~\ref{tab:parameters}. Boom lengths were evaluated on a grid of 0.25 m, so the reported optimum is the best sampled length rather than a continuous minimum.

\begin{figure}[t]
    \centering
    \includegraphics[width=\linewidth]{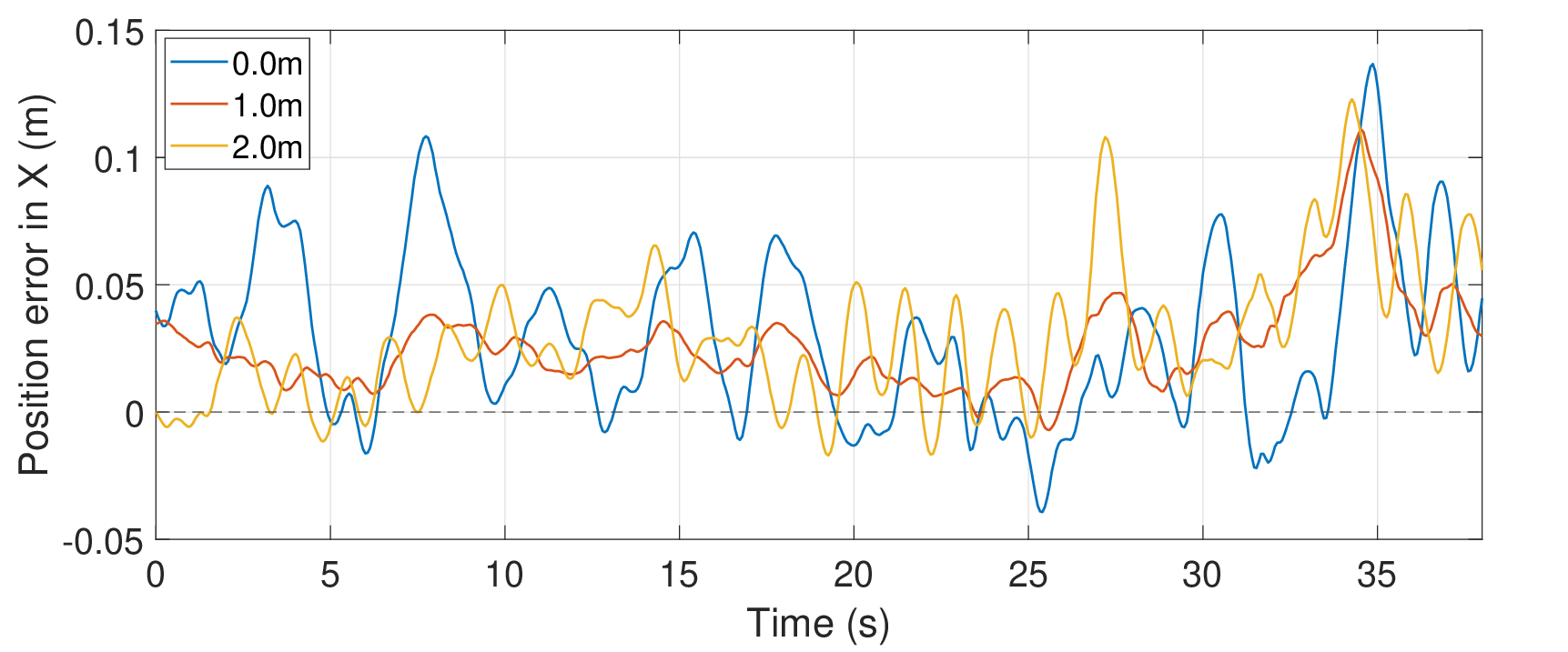}
    \caption{Position error in $x$ for a simulated flight with three different boom lengths. }
    \label{fig:example}
\end{figure}

As the dominant wind loading is along $x$, Fig.~\ref{fig:example} shows the position error in $x$ for three boom lengths. All three follow the same general displacement trend, but the 0.0 m and 2.0 m booms show larger amplitudes and faster transitions. Specifically, at 0.0 the drone is very reactive while for a 2.0 m boom the added inertia leads to oscillatory behavior. To evaluate the impact of boom length, we use the standard deviation of the position error. The mean wind pushes the drone to a small steady offset, which is similar for every boom length. The boom length instead changes how much the drone moves around that offset. The standard deviation captures exactly this movement, and grows both when the drone reacts too late with a short boom and when it oscillates with a long one. 


\begin{figure}[t]
    \centering
    \includegraphics[width=\linewidth]{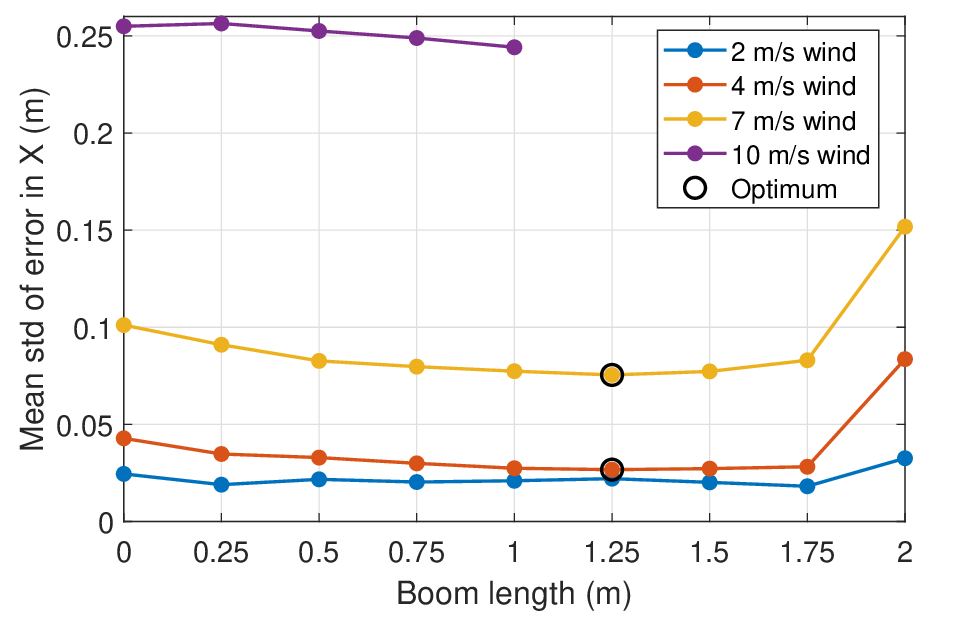}
    \caption{Standard deviation of the position error in $x$ against boom length, for four mean wind speeds. Longer booms give more preview time but add inertia, producing an optimum at 1.25 m for 4 and 7 m/s (circled). At 2 m/s the drone is fast enough that boom length has little effect, and at 10 m/s the preview time is too short to help much.}
    \label{fig:std_sim}
\end{figure}

Figure \ref{fig:std_sim} gives an overview of the mean standard deviation of the position error over the range of boom lengths and wind speeds evaluated. Boom length translates to a preview time through~\eqref{eq:timedelay}, so the same boom buys less time as the wind speed increases. For 4 m/s and 7 m/s, the standard deviation drops with boom length until the added inertia of the boom begins to penalize performance, so an optimum boom length exists for our platform. Beyond 2 m, the drone is no longer able to remain in flight. At 2 m/s, the drone is fast enough relative to the incoming gusts that the boom length has no notable effect. At 10 m/s the preview window is too short for the additional length to help, and beyond 1 m the drone becomes unstable.

\begin{figure}[t]
    \centering
    \includegraphics[width=\linewidth]{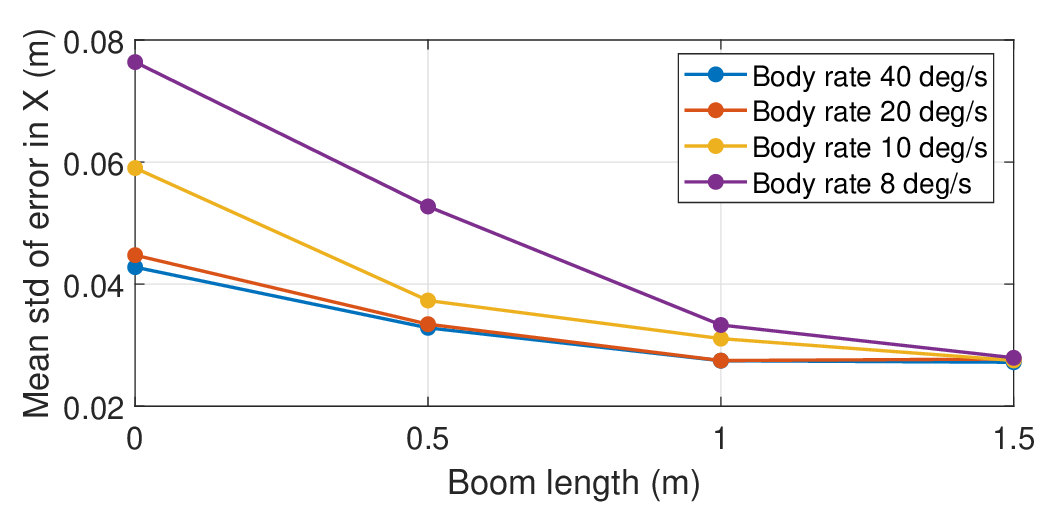}
    \caption{Performance improvement with boom length at 4 m/s wind for an artificially slowed drone, obtained by limiting the maximum body rate in the MPC (deg/s). Slower platforms benefit more from preview. }
    \label{fig:pitch}
\end{figure}

The benefit of additional preview time also depends on how quickly the drone is able to respond. We demonstrate this in Fig.~\ref{fig:pitch} by simulating a 4 m/s wind while tightening the constraint on the maximum body rate in the MPC. The tilt needed to reject a given gust is set by the disturbance rather than by the preview, so a longer boom does not change the required attitude but increases the time available to reach it. Anticipation is therefore worth more the slower the drone is to respond. Once the boom is long enough that the required rate falls below the bound, the constraint is no longer active and all rate limits converge to the same performance.

\subsection{Hardware experiments}
The effectiveness of the proposed method is demonstrated by means of two experiments. Firstly, in controlled indoor conditions, we compare against a baseline and a wind-unaware controller, and confirm the boom length trend observed in simulation over the range that can be flown safely. Secondly, we test outdoors in real wind to show deployability and the impact of a single wind-aligned sensor.

\begin{figure}[t]
    \centering
    \includegraphics[width=\linewidth]{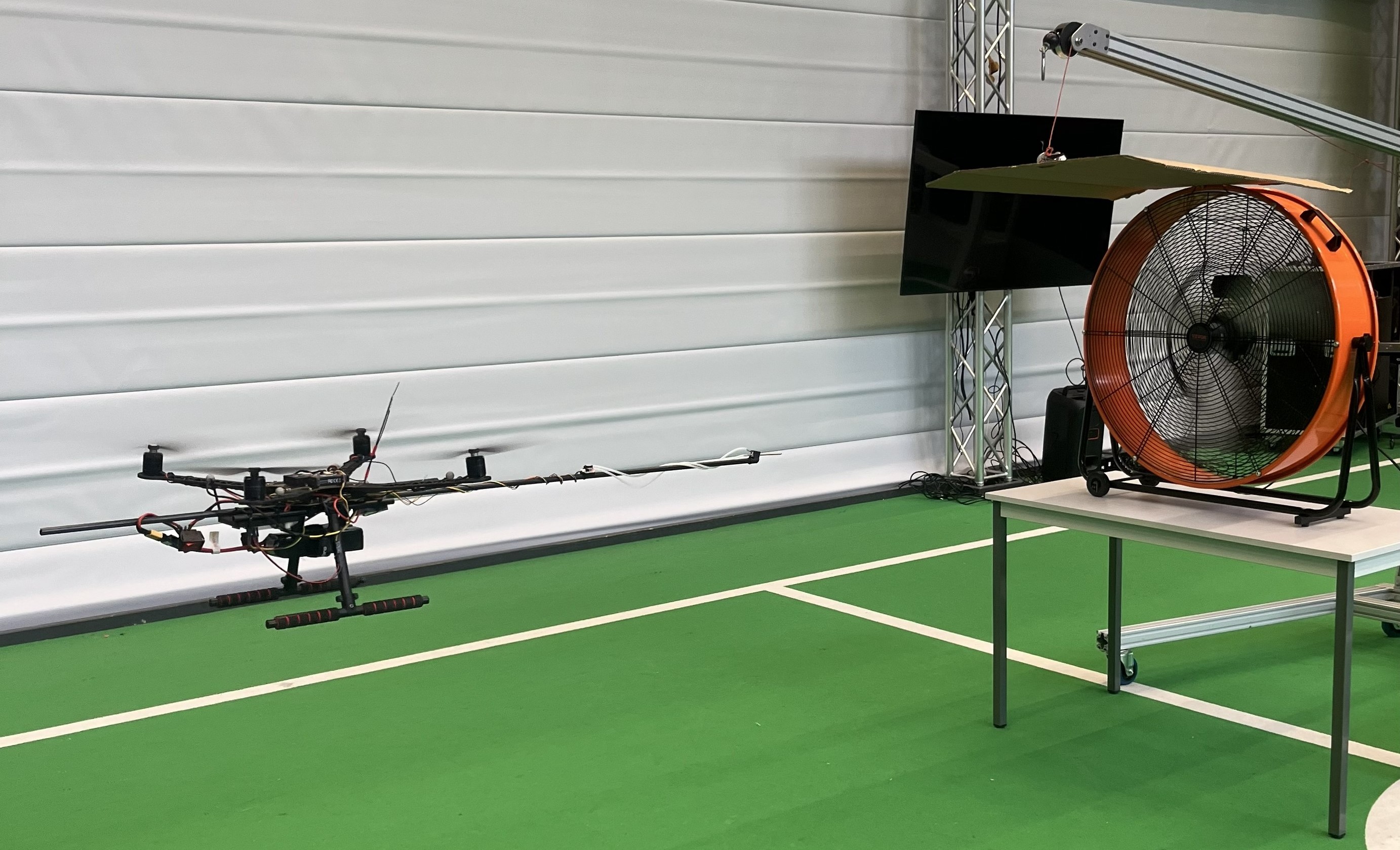}
    \caption{Indoor experimental setup where the drone hovers in front of a fan, with the boom-mounted pitot tube pointing into the flow. A sheet in front of the fan is opened rapidly to generate a repeatable gust.}
    \label{fig:experiment}
\end{figure}

Figure \ref{fig:experiment} shows the indoor setup. The wind is generated using an industrial fan (13000 CFM) placed 2.5 m from the hover position, operated at a high (approximately 4 m/s) or low (approximately 2 m/s) setting. As the fan requires a long ramp up time, a sheet is mounted in front of it that can be opened rapidly to induce a sudden gust. The fan does not produce a spatially uniform profile, which contributes to the run-to-run spread reported below. In each test, the drone hovers at 1 m altitude. Every setting is repeated 10 times, with the runs aligned on gust onset before averaging. These tests consist of isolated gusts, so there is no mean wind speed to which the estimator can converge. We therefore use a short averaging window $T_{\mathrm{MA}}$ of 0.1 s in the time delay estimate \eqref{eq:timedelay}.

We use this setup to compare the proposed approach against two controllers. The first is the widely used PX4 position mode, which we take as baseline. The second is a wind-unaware MPC, which uses the same architecture and parameters but with the sensor measurement fixed to zero. The difference between the wind-aware and wind-unaware controllers therefore isolates the benefit of the wind measurement.

\begin{figure}[t]
    \centering
    \includegraphics[width=\linewidth]{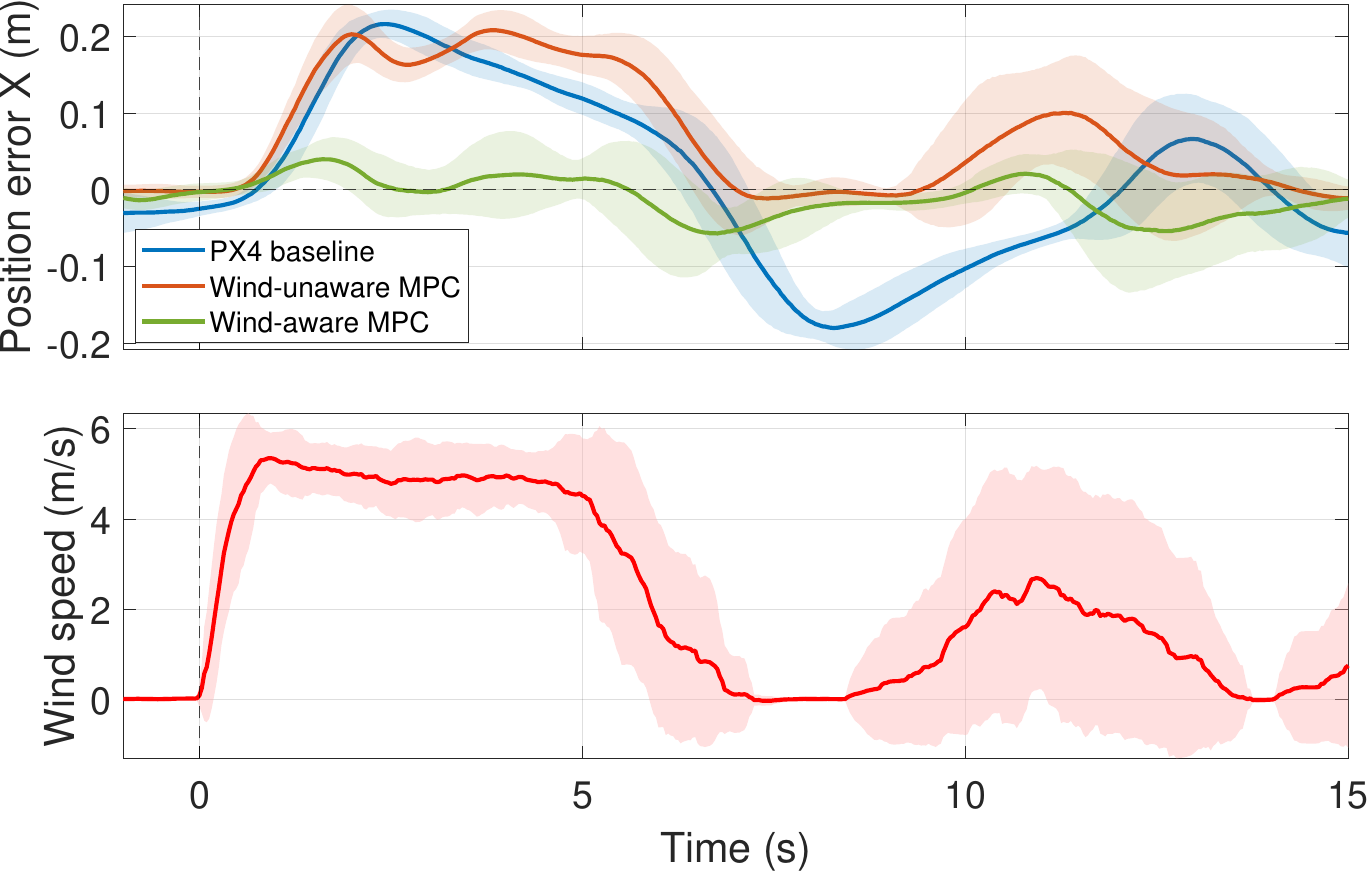}
    \caption{(Top) Mean position error (solid) with one standard deviation (shaded) for each controller. RMSE over the window is 0.144 m for the PX4 baseline, 0.120 m for the wind-unaware MPC, and 0.048 m for the wind-aware MPC. (Bottom) Mean wind speed at the boom, pooled across all runs.}
    \label{fig:indoor_compare}
\end{figure}

Figure~\ref{fig:indoor_compare} shows the mean position error in $X$ with one standard deviation (top), together with the gust the drone was subject to (bottom). The wind-aware MPC keeps the displacement well below that of the other two controllers. The wind-unaware controller is initially displaced similarly to the baseline, but does not overshoot on return. Over the window shown, the wind-aware MPC reduces the RMSE from 0.144 m for the baseline and 0.120 m for the wind-unaware MPC to 0.048 m, a reduction of 66 and 60 percent respectively.

\begin{figure}[t]
    \centering
    \includegraphics[width=\linewidth]{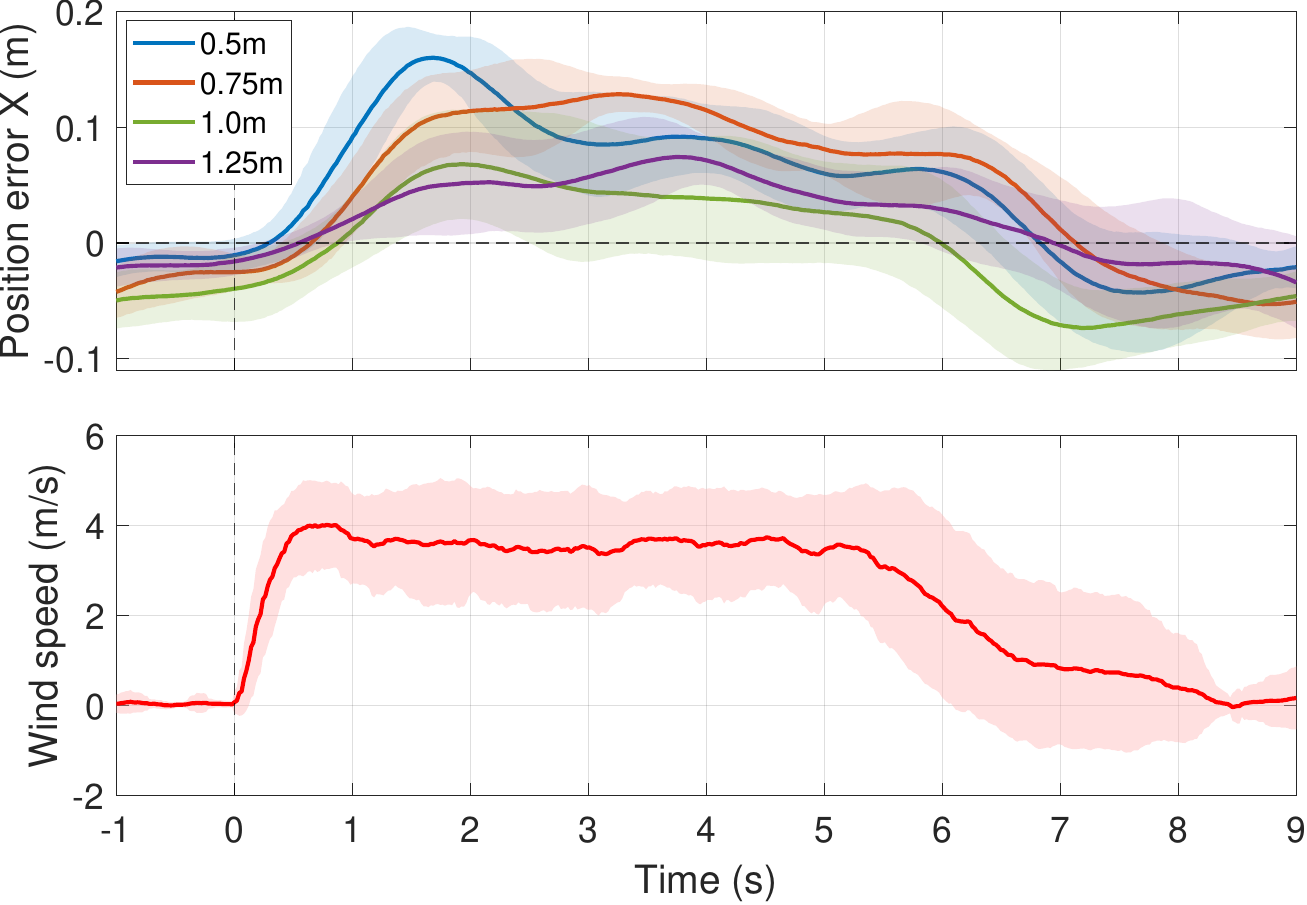}
    \caption{(Top) Mean position error (solid) with one standard deviation (shaded) for four different boom lengths. (Bottom) Mean wind speed at the boom, pooled across all runs.}
    \label{fig:length}
\end{figure}

Figure~\ref{fig:length} and Table~\ref{tab:results}, report the results of a boom length comparison. It was experimentally determined that the boom had to be mounted at least 0.4 m away to minimize the effect of propeller downwash on the sensor readings. As in simulation, performance improves with boom length for the 4 m/s gust, while no significant effect is observed at 2 m/s. With a boom of 1.5 m the flight performance was no longer considered safe, which is consistent with the degradation the simulation predicts past the optimum.

\begin{table}[t]
    \centering
    \caption{Hardware performance for different boom lengths at two wind speeds. Values are the mean RMSE over 10 runs, with the standard deviation in parentheses.}
    \begin{tabular}{c|cc} \toprule
        \textbf{Boom length [m]} & \textbf{RMSE [m] (2 m/s)} & \textbf{RMSE [m] (4 m/s)} \\ \midrule
        0.5  & 0.0593 (0.010) & 0.0836 (0.009) \\
        0.75 & 0.0602 (0.008) & 0.0722 (0.009) \\
        1.0  & 0.0569 (0.010) & 0.0652 (0.017) \\
        1.25 & 0.0552 (0.006) & \textbf{0.0538} (0.009) \\ \bottomrule
    \end{tabular}
    \label{tab:results}
\end{table}

\begin{figure}[t]
    \centering
    \includegraphics[width=\linewidth]{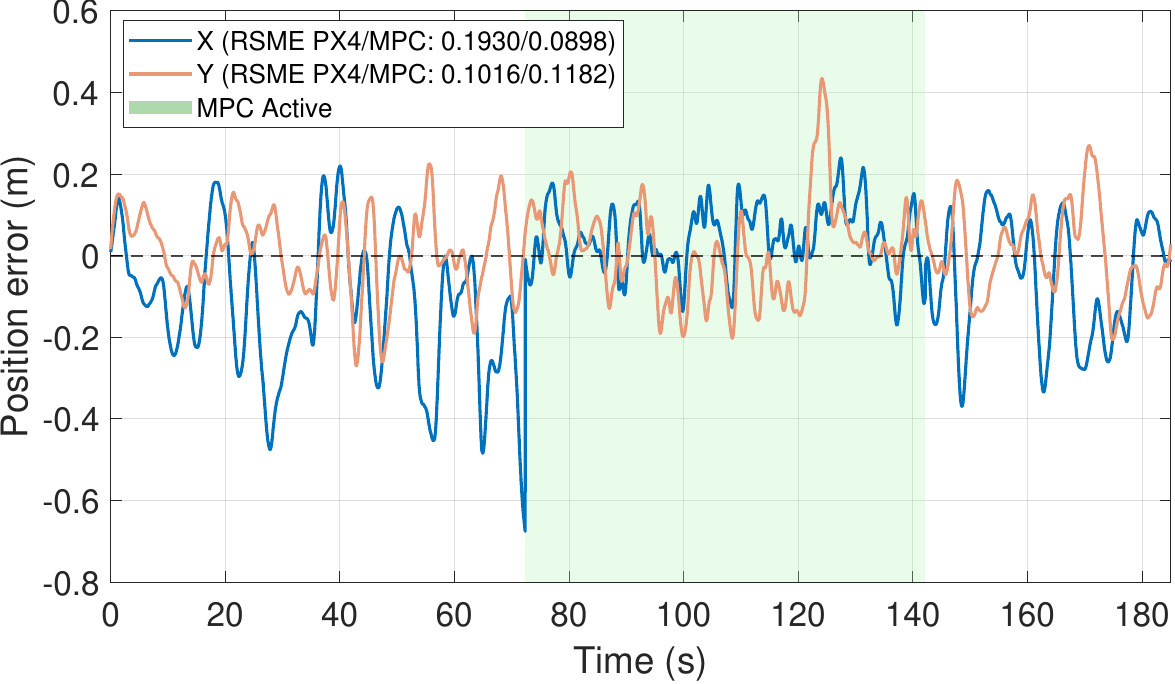}
    \caption{Position error for the outdoor experiments. RMSE along $x$, the direction aligned with the boom, improves by 54 percent, while $y$ stays similar. The setpoint changes to the current position when changing controllers.}
    \label{fig:outdoor_result}
\end{figure}

The drone is flown outdoors (Fig.~\ref{fig:outdoor}) to demonstrate the benefit of the proposed method in real-world conditions. The prevailing wind direction was measured with a ground-based ultrasonic anemometer prior to flight. The drone was then positioned with the sensor pointing in this direction, after which the GPS heading reference was initialized so that the mean flow acts along $x$. Figure~\ref{fig:outdoor_result} shows the position error in $x$ and $y$, with shading indicating when the wind-aware MPC is active and when PX4 position mode is active. During the flight, the mean wind speed was 4 m/s with gusts up to 7.5 m/s. The wind alignment is visible in the errors themselves, as the PX4 RMSE in $x$ is 47 percent larger than in $y$. Switching controllers has little effect in $y$, while in $x$, the direction in which the preview acts, the RMSE drops by 54 percent. Hovering performance can therefore be improved substantially by aligning the single sensor with the dominant disturbance direction.

\subsection{Discussion}
The results show that onboard wind-preview MPC substantially reduces hovering error in simulation and in both indoor and outdoor hardware experiments. The optimum in boom length that emerges from our results follows from two main competing effects. A larger sampling distance gives the drone more time to react, while the added inertia degrades the attitude response. A higher wind speed shifts this balance twice, since it shortens the preview time obtained from a given length and increases the force to be rejected which makes the impact of the added inertia more significant. The optimum is therefore a property of the boom length, wind speed, and the platform together. The 1.25 m found here applies to the S500, and a heavier or less agile platform is expected to favour a longer boom, as suggested by Fig.~\ref{fig:pitch}.

The setup assumes that a dominant wind direction exists and remains stable during flight to allow for alignment. This holds in open terrain and offshore, but not in strongly obstructed environments where the flow direction changes rapidly. Multiple booms, multiple sensors, or multi-hole probes, which widen the acceptance cone from roughly $\pm$20$^\circ$ to $\pm$45$^\circ$~\cite{fuertes2019MultirotorUAVBasedPlatform}, would relax the alignment requirement. 


The 0.72 m/s RMSE and the unreliable readings below roughly 1 m/s are the price of a 5 g, low-cost probe, yet the controller still significantly improves hovering performance. Preview therefore appears tolerant to measurement error, consistent with the robustness to timing and magnitude errors reported for LiDAR-based preview MPC~\cite{mendez2023WindPreviewBasedModel}. Our study, therefore, confirms that a simple sensor is sufficient for preview, which keeps the added hardware cost and mass low enough to be carried alongside the actual mission payload.

\section{CONCLUSION}\label{Se:Co}
In this work we propose to improve drone hover performance using fully onboard wind preview, obtained from a low-cost, low-weight pitot-static sensor mounted on a boom ahead of the airframe. A nonlinear MPC uses this preview to generate anticipatory actions. In simulation we characterize the trade-off between the preview gained from a longer boom and the flight performance lost to its added inertia. We show that an optimum exists and that it varies with the wind condition and the responsiveness of the drone. Indoor experiments confirm this trend and show a reduction in hover RMSE of 66 percent against a PX4 baseline and 60 percent against an otherwise identical wind-unaware MPC. Outdoors, aligning the boom with the prevailing wind reduces the RMSE along that direction by 54 percent compared to the baseline controller.

In future work, we intend to integrate the sensor on a radiographic inspection drone, where a gimbal decouples the X-ray equipment from the airframe. The drone can then be aligned with the wind while the equipment remains on target, so the improved hover accuracy directly translates into improved image quality.

\bibliographystyle{IEEEtran}
\bibliography{references}

\end{document}